\documentclass[10pt,a4paper,twocolumn]{article}

\usepackage{amsmath,amssymb,amsfonts}
\usepackage{algorithm}
\usepackage{algpseudocode}
\usepackage{booktabs}
\usepackage{graphicx}
\usepackage{svg}
\usepackage{subcaption}
\usepackage{hyperref}
\usepackage{booktabs}
\usepackage[dvipsnames]{xcolor}

\usepackage{float}
\usepackage{stfloats}
\usepackage{cuted}

\usepackage[margin=2.5cm]{geometry} 
\usepackage{CJKutf8}
\usepackage{indentfirst}

\usepackage{titlesec}
\titleformat{\section}
{\bfseries\fontsize{12pt}{12pt}\selectfont}
{\thesection.\quad}
{0pt}
{}
\titleformat{\subsection}
{\bfseries\fontsize{11pt}{11pt}\selectfont}
{\thesubsection\quad}
{0pt}
{}

\usepackage{graphicx}
\usepackage{caption}
\begin{document}
\begin{CJK*}{UTF8}{bkai}

\renewenvironment{abstract}
  {\begin{center}\bfseries {\bfseries\fontsize{12pt}{14pt}\selectfont 摘要}\end{center}}

\renewcommand{\refname}{{\bfseries\fontsize{12pt}{14pt}\selectfont 參考文獻}}

\twocolumn[
\begin{center}
  {\bfseries\fontsize{12pt}{15pt}\selectfont 以逸待勞-強化學習訓練一動不如一靜}\par\vspace{0.2cm}
  {\bfseries\fontsize{12pt}{12pt}\selectfont Constant in an Ever-Changing World}\par\vspace{0.5cm}
  {\fontsize{12pt}{12pt}\selectfont 吳建中, 林俊成, 黃月華, 廖容佐}\par\vspace{0.1cm}
  
  {\fontsize{12pt}{12pt}\selectfont 輔仁大學資訊工程學系}\par\vspace{0.065cm}
  
  {\fontsize{12pt}{12pt}\selectfont E-mail: andywu.academic@gmail.com, \ cclin@csie.fju.edu.tw,
 \\ \hspace{17pt} yhhuang@csie.fju.edu.tw, \ rtliaw@csie.fju.edu.tw}\par\vspace{1em}

 {\color{CarnationPink} \fontsize{12.5pt}{12pt}\selectfont \href{https://github.com/AndyWu101/CIC}{https://github.com/AndyWu101/CIC}}\\
\end{center}
]

\begin{abstract}
\noindent
強化學習的訓練過程常伴隨劇烈的震蕩。導致演算法的不穩定性與性能下降。
本文提出了一種以逸待勞的架構(CIC)，增強演算法的穩定性以提升效能。
CIC有代表策略以及當前策略。CIC不盲目變動代表策略，而是有選擇的在當前策略更優時更新代表策略。
CIC使用一種自適應調整的機制，使代表策略與當前策略共同幫助critic訓練。
我們分別在MuJoCo的5個環境上測試了CIC的表現。
結果顯示CIC可以在不增加計算成本的情況下提升傳統演算法的性能。

\noindent {\bfseries 關鍵字}: 強化學習、Actor-Critic、連續控制任務

\end{abstract}

\begin{figure*}[!ht]
    \centering
    \includegraphics[width=\textwidth]{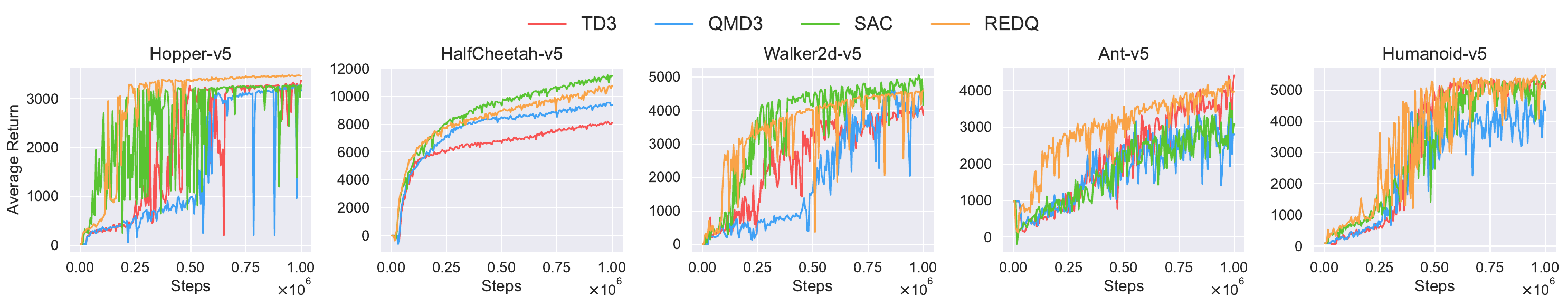}
    \caption{Actor-Critic 演算法的不穩定性}
    \label{fig:RL_instability}
\end{figure*}


\begin{figure*}[b]
    \centering
    \includegraphics[width=0.8\textwidth]{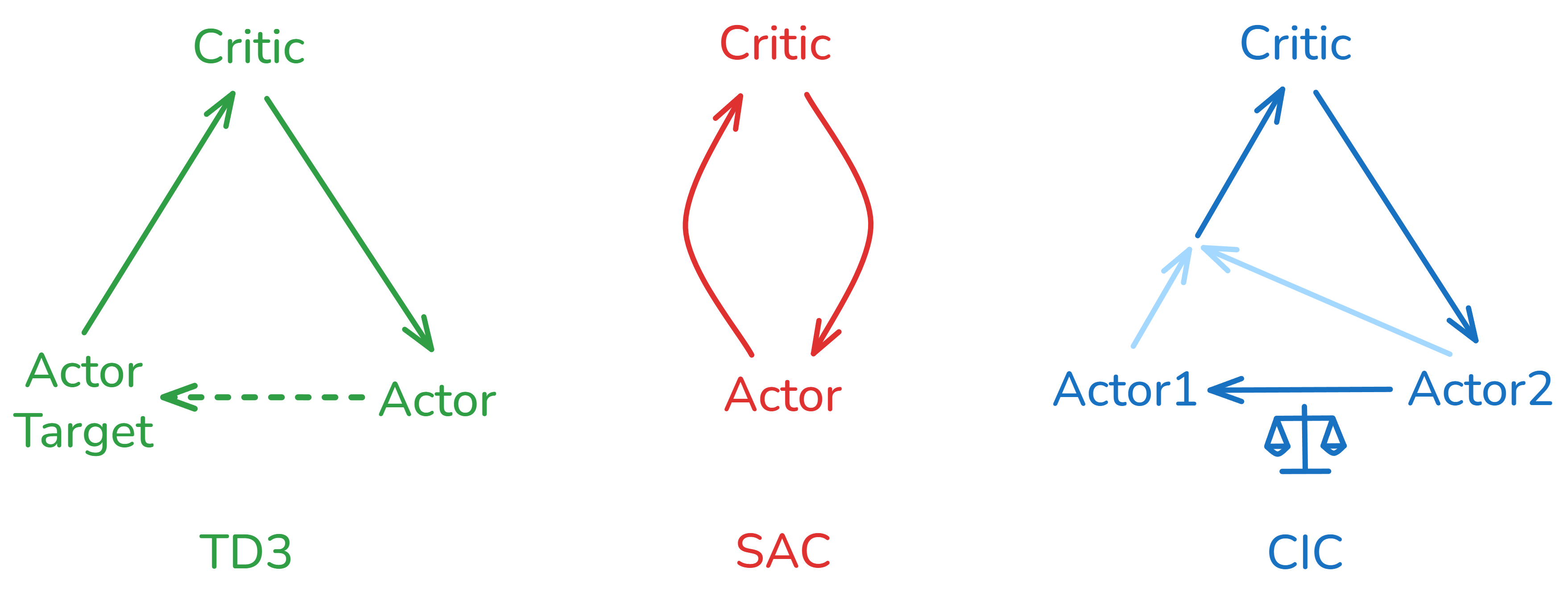}
    \caption{CIC 架構圖}
    \label{fig:CIC_architecture}
\end{figure*}

\begin{algorithm*}[ht!]
    \caption{CIC}\label{alg:1}
    \begin{algorithmic}[1]

        \State Initialize actor1 $\pi_{\phi_{1}}$ with random parameters $\phi_{1}$
        \State Initialize actor2 $\pi_{\phi_{2}} \gets \pi_{\phi_{1}}$
        \State Initialize critic networks $Q_{\theta_{i}}$ with random parameters $\theta_{i}$ for $i \in \{1 \cdots q\}$
        \State Initialize critic target networks $\theta^{\prime}_{i} \gets \theta_{i}$ for $i \in \{1 \cdots q\}$
        \State $t \gets$ Initialize replay buffer $\mathcal{R}$
        \State $\lambda \gets 0$
        \State Fill lambda buffer $\Lambda$ up with ($\lambda, -\infty$)
        \newline

        \While{$t < T$}
            \State $score_{1}, steps_{1} \gets$ Evaluate($\pi_{\phi_{1}}, \mathcal{R}$)  \Comment{Store transitions $(s, a, r, s^{\prime})$ in $\mathcal{R}$}
            \State $score_{2}, steps_{2} \gets$ Evaluate($\pi_{\phi_{2}}, \mathcal{R}$)  \Comment{Store transitions $(s, a, r, s^{\prime})$ in $\mathcal{R}$}
            \State $\pi^{\mathcal{S}}_{\phi_{1}} \gets \pi^{\mathcal{S}}_{\phi_{1}} \cup score_{1}$
            \State $\pi^{\mathcal{S}}_{\phi_{2}} \gets score_{2}$
            \State $\Delta t \gets steps_{1} + steps_{2}$
            \State Replace the oldest pair of $\Lambda$ with $(\lambda, score_{2})$
            \newline

            \If{$\pi^{\bar{\mathcal{S}}}_{\phi_{1}} < \pi^{\bar{\mathcal{S}}}_{\phi_{2}}$}
                \State $\pi_{\phi_{1}} \gets \pi_{\phi_{2}}$
            \EndIf
            \newline

            \State $\lambda \gets \text{mean}\{\lambda_{i} \mid (\lambda_{i}, score_{i}) \in \text{TopHalf}_{score}(\Lambda)\}$
            \State $\lambda \gets \text{clip}(\lambda + \epsilon, 0, 1), \ \epsilon \sim \mathcal{N}(0, \sigma)$

            \For{$i \gets 1$ to $\Delta t$}
                \State $\mathcal{B} \gets$ Sample a mini-batch of $N$ transitions $(s, a, r, s^{\prime})$ from $\mathcal{R}$
                \For{$j \gets 1$ to $N$}
                    \State $(s, a, r, s^{\prime}) \gets \mathcal{B}_{j}$
                    \If{$j \le \lfloor N \cdot \lambda \rfloor$}
                        \State $a^{\prime} \gets \pi_{\phi_{1}}(s^{\prime})$
                    \Else
                        \State $a^{\prime} \gets \pi_{\phi_{2}}(s^{\prime})$
                    \EndIf
                    \State $\mathcal{B}_{j} \gets (s, a, r, s^{\prime}, a^{\prime})$
                \EndFor
                \State Train critics $Q_{\theta_{i \cdots q}}$ and actor2 $\pi_{\phi_{2}}$ by $\mathcal{B}$
                \State Update critic targets $\theta^{\prime}_{i} \gets \tau \theta_{i} + (1 - \tau)\theta^{\prime}_{i}$
            \EndFor
            \State $t \gets t + \Delta t$
        \EndWhile
        \State \Return $\pi_{\phi_{1}}$

    \end{algorithmic}
\end{algorithm*}

\section{緒論}

強化學習 (Reinforcement Learning) 被研究於搜尋遊戲最佳策略、機械
控制、訓練語言模型等方面並展現強化學習的應用價值。依照算法性質，
強化學習可分為以下兩種類別:
1) 基於價值的方法 (Value-based Approach)
2) 基於策略的方法 (Policy-based Approach)。
基於策略的方法中 Actor-Critic 是一種在連續控制任務中非常有效的架構。
本次研究主要聚焦於 Actor-Critic 方法。

在研究中我們發現，Actor-Critic 架構中的 actor 性能在訓練中有時候會突然大幅下降。
然而傳統 Actor-Critic 架構中 actor 和 critic 互為倚仗，如果某方性能下降，可能導致惡性循環，加劇演算法的不穩定性。

為了解決這個問題，本文提出了一種「以逸待勞」的新架構 (CIC)，其在不增加計算成本的情況下，
通過設計打斷惡性循環，增強演算法穩定性。
實驗結果顯示 CIC 在4種演算法上都有一定程度的性能提升。

\section{準備工作}

為了系統化地描述強化學習的過程，通常將強化學習問題建模為馬可夫決策過程(MDP\cite{sutton1998_Sutton_RL})。MDP提供了一個數學框架，透過5元組 $< \mathcal{S}, \mathcal{A}, \mathcal{R}, \mathcal{P}, \gamma >$定義，其中:
\begin{itemize}
    \item $\mathcal{S}$是狀態空間
    \item $\mathcal{A}$是動作空間
    \item $\mathcal{R}(s,a) = \mathbb{E} \bigl[r_{t} \mid s_t=s,a_t=a \bigr]$ 是回饋函數
    \item $\mathcal{P}(s_{t+ 1} \mid s_t, a_t)$是指從狀態$s_t$採取動作$a_t$後環境狀態轉移成$s_{t + 1}$的機率分布
    \item $\gamma \in [0,1]$是折扣因子，代表對於未來回報的重視程度
\end{itemize}

在強化學習中，actor 在每個離散時間步$t$接收環境給予的狀態$s_t \in \mathcal{S}$並依據其策略$\pi$選擇動作$a_t \sim \pi(\cdot \mid s_t)$，獲得環境的回饋$\mathcal{R}(s_t, a_t)$。actor的目標是學習最佳策略獲得最大的
總折扣回報
$$R_t = \sum_{i = t} ^ T \gamma^{i - t} \mathcal{R}(s_{t}, a_{t})$$
對於給定$\pi$，可以定義狀態-動作價值函數(或稱Q函數)
$$Q^{\pi}(s_{t},a_{t}) = \mathbb{E} \Big[ R_t \ \Big{|} \ s_{t},a_{t} \Big]$$
同時Q函數滿足Bellman Expectation Equation\cite{sutton1998_Sutton_RL}:
$$Q^\pi(s_t,a_t) = \mathbb{E} \Big[ \mathcal{R}(s_t, a_t) + \gamma \mathbb{E}[Q^\pi(s_{t+1},a_{t+1})] \Big]$$
Actor-Critic的方法中:
\\

Critic的損失函數定義為最小化
$$
    J_{ Q^{\pi} } = \mathbb{E}\Big[ Q^{\pi}(s_t, a_t) - y_{t} \Big],
$$
$$
    y_{t} = \mathcal{R}(s_{t}, a_{t}) + \gamma Q^{\pi}(s_{t+1}, a_{t+1})
$$
\\
\indent Actor的目標函數定義為最小化
$$
J_{\pi} = \mathbb{E}\Big[-Q^{\pi}(s_t, a_t ) \ \Big{|} \  a_t \sim \pi(\cdot \mid s_t)\Big]
$$

\section{相關研究}

在連續控制任務中著名算法 DDPG\cite{lillicrap2015_DDPG} 採用 Actor-Critic 方法，使用單一 critic 估計 Q 值以及單一確定性策略的 actor，並且提出軟更新(sort update)的概念，後續成為強化學習算法中常見的技巧。因此DDPG在連續控制任務中具有承先啟後的意義。

然而DDPG由於訓練時Q值高估導致的不穩定導致收斂不佳。為解決此問題，TD3\cite{fujimoto2018_TD3}引入兩個 critic 取最小值以估計目標Q值，減緩Q值高估問題，同時為 actor 引入actor target以及延遲更新，增加訓練過程的穩定性。

TD3由於估計Q值時使用最小值操作，導致Q值低估的現象抑制了actor的探索，為了改善此現象，QMD3\cite{wei2022_QMD3}提出使用$N$個critic (推薦$N = 4$)，將每個critic估計的Q值排序後取第$\lfloor \frac{N}{2} \rfloor$個Q值作為最終估計，能夠緩解Q值高估與低估的發生。

與TD3同一時期SAC\cite{haarnoja2018_SAC}一樣使用兩個critic取最小值以估計目標Q值，不同的是SAC採用隨機策略的actor以及基於soft Q函數 \cite{haarnoja2017_Soft_Q_Learning} 訓練critic，soft Q函數除了原有的環境回饋之外引入動作的熵，如此critic擁有引導actor嘗試更多樣動作的能力。

後續的研究中，REDQ\cite{chen2021_REDQ}在SAC的基礎上使用$N$個critic (推薦$N=10$)。具體而言，在訓練critic時，REDQ每次從$N$個critic中隨機抽選2個計算目標Q值；在訓練actor時，REDQ使用$N$個critic的平均梯度訓練actor。以此增強穩定性，進一步提升SAC的效能。

\section{穩定性分析}
Figure \ref{fig:RL_instability} 中我們展示了4種Actor-Critic演算法在5個環境中單次訓練的結果，可以看到除了HalfCheetah-v5以外，其他環境的訓練過程皆相當震蕩，有時甚至會從接近滿分瞬間掉到接近0分。

\section{方法}
本研究提出了一個新穎的Actor-Critic架構，Figure \ref{fig:CIC_architecture}展示了CIC和傳統演算法的架構差異。
CIC通過以逸待勞以及自適應調整$\lambda$機制，改善了傳統演算法的穩定性。
Algorithm \ref{alg:1}描述了演算法流程，詳細介紹如下。

\subsection{以逸待勞}

在傳統方法中，actor或actor target都會無條件的進行改變，哪怕改變會導致actor或actor target的性能下降。
為了解決這個問題，CIC的架構有兩個actor。actor1為高分actor，其不會受到訓練或任何改變，用以維持穩定性；actor2為接受訓練的actor，用以探索並超越actor1，使演算法進步。

為了獲知 actor 的性能，我們將訓練流程從互動一步訓練一步，改成互動一局訓練一局。
由於actor1不會改變，所以我們持續紀錄其歷史得分；然而actor2每輪都會受到梯度訓練並改變參數，所以只使用當前得分。
當actor2的分數高於actor1，則actor2會成為新的actor1。

在設計中我們認為actor1的分數將於一定局數後逐漸收斂，同時為了避免失去動作多樣性，所以我們限制actor1與環境互動的最高局數為10局。
另外由於環境和策略本身的隨機性，有時需要玩更多局以獲知actor的真正性能。如果只玩一局，容易因為運氣獲得高分導致actor2的得分高於actor1。
因此CIC使用參數 $\kappa$ 允許actor2起始評估超過一局，以多局分數作為基準，確保actor2分數的可信度。

\begin{figure*}[!b]
    \begin{subfigure}{\textwidth}
        \centering
        \includegraphics[width=\textwidth]{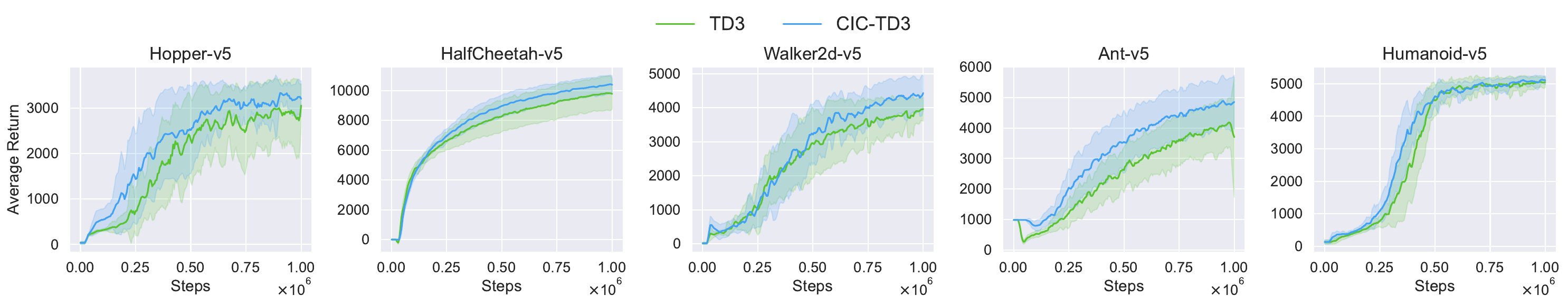}
        \label{fig:CIC-TD3}
    \end{subfigure}

    \begin{subfigure}{\textwidth}
        \centering
        \includegraphics[width=\textwidth]{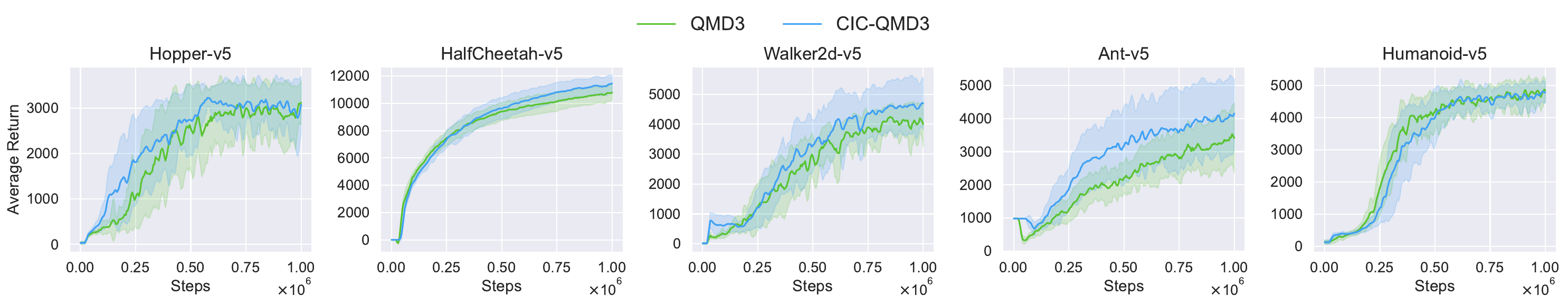}
        \label{fig:CIC-TD3}
    \end{subfigure}

    \begin{subfigure}{\textwidth}
        \centering
        \includegraphics[width=\textwidth]{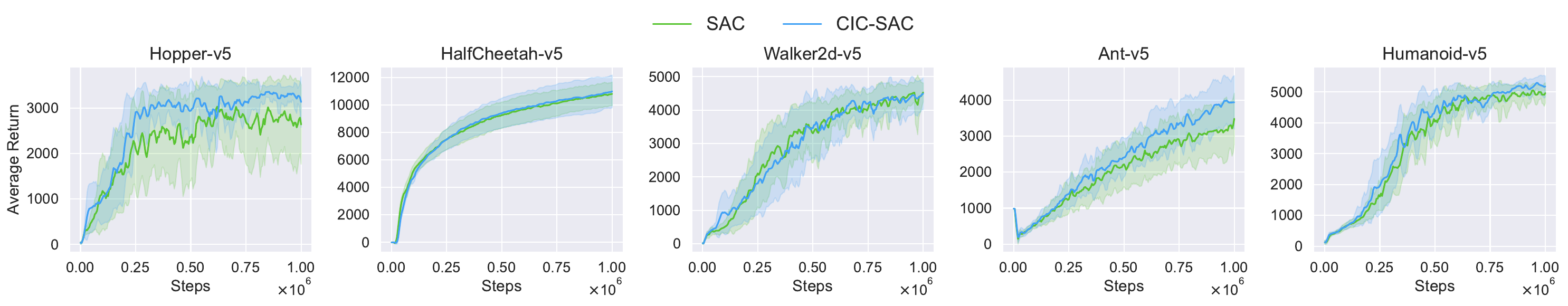}
        \label{fig:CIC-SAC}
    \end{subfigure}

    \begin{subfigure}{\textwidth}
        \centering
        \includegraphics[width=\textwidth]{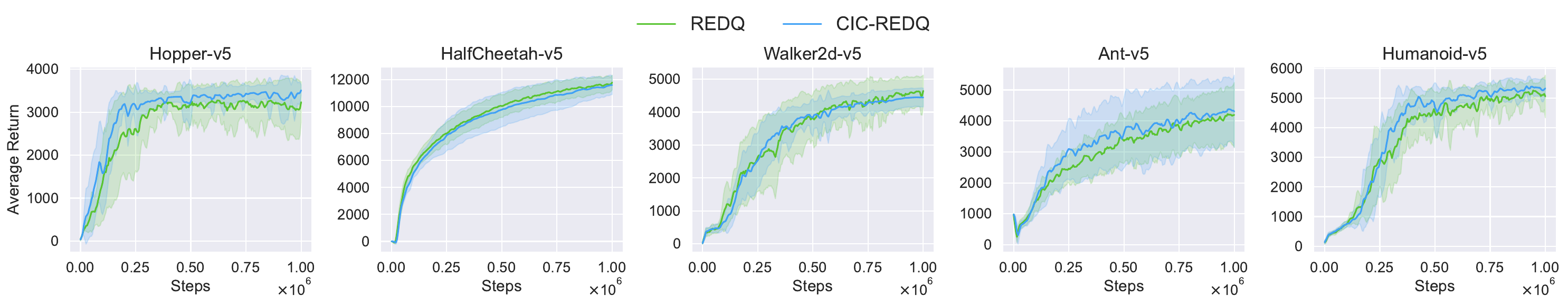}
        \label{fig:CIC-REDQ}
    \end{subfigure}

    \caption{CIC 對比實驗}
    \label{fig:CIC_main_exp}
\end{figure*}

\begin{figure*}[!t]
    \centering
    \includegraphics[width=\textwidth]{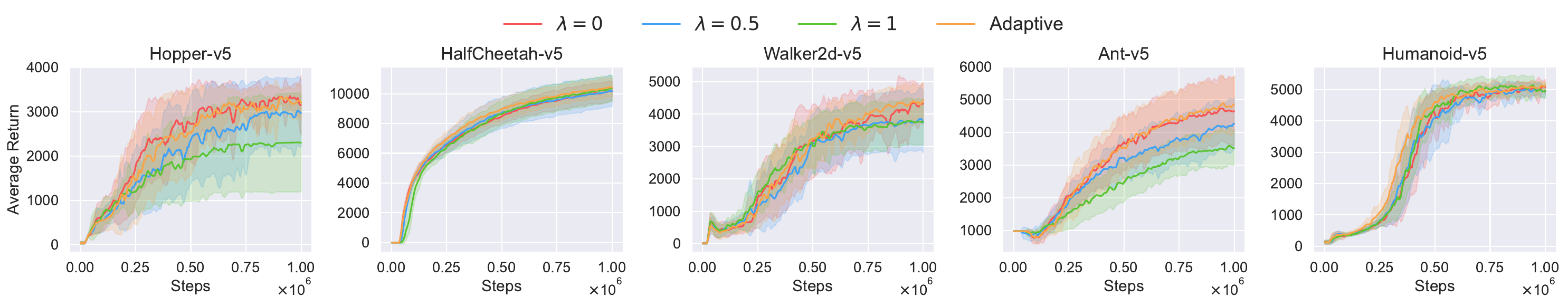}
    \caption{CIC-TD3 \ 固定$\lambda$分析實驗}
    \label{fig:CIC-TD3_fixed_lambda_exp}
\end{figure*}

\begin{figure*}[!t]
    \centering
    \includegraphics[width=\textwidth]{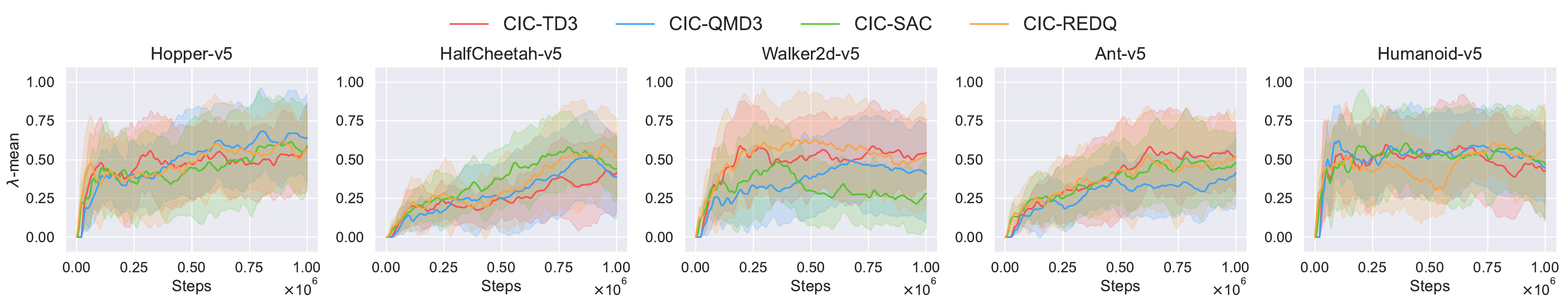}
    \caption{CIC \ $\lambda$ 隨訓練過程自適應調整變化}
    \label{fig:CIC_lambda_adaptive}
\end{figure*}

\begin{table*}[!b]
    \centering
    \caption{各演算法平均回撤}
    \label{tab:avg_drawdown}
    \resizebox{\textwidth}{!}{
        \begin{tabular}{l | cc | cc |cc | cc}
            \toprule
              & TD3 & CIC-TD3 & QMD3 & CIC-QMD3 & SAC & CIC-SAC & REDQ & CIC-REDQ \\

            \cmidrule(r){1-1} \cmidrule(lr){2-3} \cmidrule(lr){4-5} \cmidrule(lr){6-7} \cmidrule(l){8-9}   
             
             Hopper-v5      & $130 \pm 42$ & \boldmath{$66 \pm 30$} & $145 \pm 64$ & \boldmath{$48 \pm 26$} & $235 \pm 95$ & \boldmath{$88 \pm 30$} & $100 \pm 65$ & \boldmath{$31 \pm 30$} \\
             HalfCheetah-v5 & $71 \pm 28$ & \boldmath{$32 \pm 8$} & $71 \pm 19$ & \boldmath{$20 \pm 7$} & $67 \pm 22$ & \boldmath{$25 \pm 7$} & $86 \pm 20$ & \boldmath{$19 \pm 5$} \\
             Walker2d-v5    & $107 \pm 24$ & \boldmath{$91 \pm 25$} & $181 \pm 31$ & \boldmath{$85 \pm 24$} & $170 \pm 22$ & \boldmath{$125 \pm 45$} & $122 \pm 35$ & \boldmath{$34 \pm 21$} \\
             Ant-v5         & $137 \pm 23$ & \boldmath{$88 \pm 23$} & $176 \pm 21$ & \boldmath{$86 \pm 16$} & $155 \pm 25$ & \boldmath{$140 \pm 11$} & $185 \pm 31$ & \boldmath{$85 \pm 17$} \\
             Humanoid-v5    & $96 \pm 20$ & \boldmath{$65 \pm 11$} & $181 \pm 27$ & \boldmath{$75 \pm 18$} & $198 \pm 24$ & \boldmath{$113 \pm 21$} & $203 \pm 56$ & \boldmath{$74 \pm 27$} \\
            \bottomrule
        \end{tabular}
    }
\end{table*}

\subsection{$\lambda$與其自適應調整機制}

CIC不使用actor target，而是使用actor1和actor2共同幫助critic訓練，並通過一個自適應調整的係數 $\lambda$ 控制 actor1 的參與比例，這使得 critic 的訓練更穩定，從而打斷惡性循環。
具體來說，CIC訓練critic時透過$\lambda$控制一定比例的mini-batch由actor1決定$a^{\prime}$，剩餘由actor2決定。
如此critic將同時引入actor1與actor2的知識。

$\lambda$在不同環境以及不同訓練階段有不同的最佳值。因此 CIC 設計了一個機制自適應調整$\lambda$。
CIC 使用一個緩衝區$\Lambda$紀錄過去一段時間的$\lambda$與同時期actor2的分數。
每次開始訓練前，從$\Lambda$中取分數最高的一半，計算這些$\lambda$的平均值，再加上一個常態擾動探索新的$\lambda$，同時確保$0 \le \lambda \le 1$。
獲得actor2分數後，將$\Lambda$中最舊的紀錄替換為當前$(\lambda, \text{actor2 分數})$。
如此CIC能夠透過訓練過程的資訊自適應調整$\lambda$。


             

\section{實驗}

\subsection{MuJoCo}
MuJoCo\cite{6386109_MuJoCo} 全名為 Multi-Joint dynamics with Contact，是一款物理引擎，主要用於提供真實的物理模擬，適用於需要快速且精準模擬的場合。其特色在於能夠同時兼顧物理精確性與計算效率，特別針對機器人與環境之間的物理接觸進行建模與模擬。本研究採用官方最新的版本v5。
\href{https://github.com/Farama-Foundation/Gymnasium}{https://github.com/Farama-Foundation/Gymnasium}

\subsection{實驗設定}
我們在MuJoCo的5個環境上(Hopper, HalfCheetah, Walker2d, Ant, Humanoid)測試演算法性能。
我們每5000步以 actor 20局的平均性能作為基準。
學習曲線由10個seed的平均組成，陰影部分為正負一倍標準差，曲線經過平滑化。
Table \ref{tab:hyperparameter_setting} 展示了實驗的超參數設定。

\subsection{實驗結果}
Figure \ref{fig:CIC_main_exp} 顯示，
CIC-TD3在Hopper-v5、HalfCheetah-v5、Walker2d-v5、Ant-v5上有明顯的效能提升，在Humanoid-v5有較快的收斂速度。
CIC-QMD3在Hopper-v5上收斂較快，在HalfCheetah-v5、Walker2d-v5、Ant-v5上有明顯的效能提升，但在Humanoid-v5上收斂較慢。
CIC-SAC在Hopper-v5、Ant-v5上有明顯的效能提升，其他環境則維持相同表現。
CIC-REDQ在Hopper-v5、Ant-v5、Humanoid-v5上效能有一定提升，在Walker2d-v5上維持相同表現，但在HalfCheetah-v5上收斂略慢。
Figure \ref{fig:RL_instability}中可見Hopper-v5是一個訓練過程特別震蕩的環境，但在Hopper-v5上CIC的標準差都比原演算法低，顯示了其增強穩定性的效果。

Figure \ref{fig:CIC-TD3_fixed_lambda_exp} 顯示，
在5個環境中自適應$\lambda$都可以獲得最佳性能，固定$\lambda=0$也有較佳表現，
但固定$\lambda=1$則會有較明顯的性能下降，代表策略相差太大時，不能完全依賴 actor1。
Figure \ref{fig:CIC_lambda_adaptive} 顯示，
大部分$\lambda$都會收斂在0.5，但相較於 Figure \ref{fig:CIC-TD3_fixed_lambda_exp}
中的固定$\lambda=0.5$，自適應$\lambda$會有較佳表現，代表自適應機制是有效且必要的。
Table \ref{tab:avg_drawdown} 顯示，
CIC在4種演算法應用於5種環境，共20種情況下，其平均回撤都有大幅下降，
進一步證明了CIC對穩定性的提升。

\begin{table*}[t]
\centering

\caption{超參數設定}
\label{tab:hyperparameter_setting}

\begin{tabular}{lcccc}

\toprule

\textbf{Hyper-parameter} & \textbf{TD3} & \textbf{QMD3} & \textbf{SAC} & \textbf{REDQ} \\

\midrule

Number of Critics ($q$) & $2$ & $4$ & $2$ & $10$ \\
Discount Factor ($\gamma$) & $0.99$ & $0.99$ & $0.99$ & $0.99$ \\
Learning Rate & $3 \cdot 10^{-4}$ & $3 \cdot 10^{-4}$ & $3 \cdot 10^{-4}$ & $3 \cdot 10^{-4}$ \\
Optimizer & Adam & Adam & Adam & Adam \\
Batch Size & $256$ & $256$ & $256$ & $256$ \\
Actor Target & \scalebox{0.8}{$\bigcirc$} & \scalebox{0.8}{$\bigcirc$} & \scalebox{1.2}{$\times$} & \scalebox{1.2}{$\times$} \\
Critic Target & \scalebox{0.8}{$\bigcirc$} & \scalebox{0.8}{$\bigcirc$} & \scalebox{0.8}{$\bigcirc$} & \scalebox{0.8}{$\bigcirc$} \\
Soft Update Ratio ($\tau$) & $5 \cdot 10^{-3}$ & $5 \cdot 10^{-3}$ & $5 \cdot 10^{-3}$ & $5 \cdot 10^{-3}$ \\
UTD Ratio & $1$ & $1$ & $1$ & $1$ \\
Delay Frequency & $2$ & $2$ & \scalebox{1.2}{$\times$} & \scalebox{1.2}{$\times$} \\
Warmup Steps & $25000$ & $25000$ & $10000$ & $5000$\\
Exploration Noise & $\mathcal{N}(0,0.1)$ & $\mathcal{N}(0,0.1)$ & \scalebox{1.2}{$\times$} & \scalebox{1.2}{$\times$} \\
Target Policy Noise & $\mathcal{N}(0,0.2)$ & $\mathcal{N}(0,0.2)$ & \scalebox{1.2}{$\times$} & \scalebox{1.2}{$\times$} \\
Policy Noise Clip & $[-0.5,0.5]$ & $[-0.5,0.5]$ & \scalebox{1.2}{$\times$} & \scalebox{1.2}{$\times$} \\
Temperature ($\alpha$) & \scalebox{1.2}{$\times$} & \scalebox{1.2}{$\times$} & Adaptive & Adaptive \\
Target Entropy & \scalebox{1.2}{$\times$} & \scalebox{1.2}{$\times$} & $-|\mathcal{A}|$ & $\{-4,-3,-2,-1\}$ \\
Log Std Clip & \scalebox{1.2}{$\times$} & \scalebox{1.2}{$\times$} & $[-20,2]$ & $[-20,2]$ \\
Ensemble Subset Size & \scalebox{1.2}{$\times$} & \scalebox{1.2}{$\times$} & \scalebox{1.2}{$\times$} & $2$ \\

\toprule

 & \textbf{CIC-TD3} & \textbf{CIC-QMD3} & \textbf{CIC-SAC} & \textbf{CIC-REDQ} \\

\midrule

Actor2 Evaluations ($\kappa$) & $1$ & $2$ & $1$ & $2$ \\
Lambda Buffer Size ($|\Lambda|$) & $10$ & $10$ & $10$ & $6$ \\
Lambda Std. ($\sigma$) & $0.1$ & $0.1$ & $0.1$ & $0.1$ \\
 
\bottomrule

\end{tabular}
\end{table*}


             

\section{結論}

我們發現強化學習訓練過程常伴隨較大的不穩定震蕩，這降低了演算法的可靠性。
因此本研究提出了CIC，利用2個功能不同的actor，配合自適應調整機制，提升了穩定性。
結果表明強化學習訓練「一動不如一靜」，以逸待勞或許是更高明的做法。
而且由於CIC的機制簡潔，因此可以輕鬆地將其添加到任何 Actor-Critic 演算法中。

%

\bibliographystyle{IEEEtran}
\bibliography{refs}  

\end{CJK*}
\end{document}